\documentclass[a4paper,11pt]{scrartcl}

\newcommand{\titlerunning}[1]{}
\newcommand{\authorrunning}[1]{}
\newcommand{\tocauthor}[1]{}
\newcommand{\toctitle}[1]{}
\newcommand{\orcidID}[1]{}
\newcommand{\institute}[1]{}
\makeatletter
\DeclareOldFontCommand{\bf}{\normalfont\bfseries}{\mathbf}
\makeatother
\usepackage[T1]{fontenc}
\usepackage{graphicx}
\usepackage{booktabs}
\usepackage[misc]{ifsym}
\usepackage{amssymb}
\usepackage{subcaption}
\usepackage{placeins}

\usepackage{url}

\usepackage{mwe}
\usepackage{amsmath}

\usepackage[ruled,vlined,linesnumbered]{algorithm2e}
\SetAlgoSkip{}
\SetAlFnt{\small\ttfamily} 
\SetNlSty{small}{}{} 
\SetAlCapFnt{\small} 
\SetAlCapNameFnt{\small}

\usepackage{tabularx}

\newcommand{\spara}[1]{\smallskip\noindent\textbf{#1}}

\usepackage{scrlayer-scrpage}
\begin{document}

\title{Measuring What Matters:\\ A Unified Evaluation Framework for GNN Explainability}

\author{
    Francesco Paolo Nerini$^{1,*}$, 
    Mirko Zaffaroni$^{2}$, 
    Paolo Baracco$^{3}$, \\ % <--- Questo impedisce di andare fuori pagina
    Gabriele Ciravegna$^{4}$, 
    Alan Perotti$^{2}$ \\[1.5em]
    \small $^{1}$ Sapienza University of Rome, Rome, Italy - \texttt{nerini@diag.uniroma1.it} \\
    \small $^{2}$ Intesa Sanpaolo AI Research, Turin, Italy \\
    \small $^{3}$ Anti Financial Crime Digital Hub, Turin, Italy \\
    \small $^{4}$ Intesa Sanpaolo Innovation Center, Turin, Italy \\
}

\date{}
\maketitle

\begin{abstract}

Graph eXplainable AI (G-XAI) is increasingly important for making Graph Neural Networks interpretable and accountable. While a growing number of explainers are available, choosing the right method and assessing the trustworthiness of its outputs remains unclear. Consistent evaluation practices and actionable guidance are still missing, hindering practical adoption. In this paper, we introduce a unified, quantitative benchmarking framework for G-XAI that requires no ground-truth assumptions. We formalize tabular explainability metrics for graph data, evaluating topological structure and node features as independent components. Our large-scale benchmarking study identifies explainers that consistently lie on the Pareto front across metric pairs and tasks, establishing robustly non-dominated solutions - while confirming that no single explainer achieves universal superiority. We distill our findings into actionable G-XAI usability guidelines to support Machine Learning practitioners in evaluating and deploying trustworthy GNN-based pipelines.\\
%\keywords{Graph Machine Learning  \and Explainable Artificial Intelligence \and Explanation Evaluation}
\end{abstract}
\vspace*{2mm}

\section{Introduction}
\label{sec:intro}

Graph Neural Networks (GNNs) have emerged as a powerful paradigm for modeling relational data, achieving state-of-the-art performance across domains such as financial transaction monitoring, social network analysis, and biological interaction networks~\cite{wu2020comprehensive,hamilton2017inductive}. However, their inherently recursive message-passing mechanisms render internal decision processes opaque. With legal frameworks such as the EU AI Act increasingly mandating transparency for high-risk AI systems, this ``black-box'' nature represents a critical barrier to deployment in regulated environments: one cannot deploy what one cannot explain.

To address this tension, the field of Graph eXplainable AI (G-XAI) has introduced numerous post-hoc techniques that augment GNN predictions with interpretable information, typically by highlighting the most relevant nodes, edges, or features~\cite{ying2019gnnexplainer,huang2022graphlime,yuan2021subgraphx}. Yet this proliferation of explainers has created a new challenge: the problem is no longer a lack of solutions, but how to navigate them: there is no consensus on what constitutes a ``good'' graph explanation, leaving ML practitioners without objective, quantitative guidance for selecting the right tool for a given pipeline. Moreover, while the literature heavily favors node and graph classification for attribution-based methods, real-world applications routinely involve link classification and regression tasks as well.

Recent evaluation efforts have either focused on tabular or image data~\cite{perotti2024explainability}, or relied on synthetic graph datasets with artificially injected ground-truth motifs~\cite{rathee2022bagel,agarwal2023graphxai,Monti2024ATA}.
However, in real scenarios the model may have internalized patterns that diverge from human-defined labels, and we argue that it is the model's behavior, not the oracle, that explanations must faithfully reflect. Furthermore, a production-ready G-XAI framework must also focus on the pragmatic properties of explanation generation time, algorithmic reproducibility and human comprehensibility.\\

This work is guided by the following Research Questions:
\begin{description}
\item[\textbf{RQ1}] How can core explanation properties be rigorously operationalized for graph-structured data, accounting for both topological and feature-level attributions, without relying on synthetic ground-truth motifs?
\item[\textbf{RQ2}] How do state-of-the-art graph explainers compare across different GraphML tasks, and what trade-offs emerge between core explanation properties?
\item[\textbf{RQ3}] Can benchmarking results transfer to operational guidelines to guide practitioners in explainer selection?
\end{description}

\noindent
To answer these questions, we introduce a unified, quantitative framework for benchmarking G-XAI methods across the full spectrum of GraphML tasks.
Our main contributions are: (i) {\bf a quantitative, ground-truth-free evaluation framework for graphs}, where we adapt and extend three tabular metrics - \textit{Stability}, \textit{Pertinence}, and \textit{Effective Compactness}, in addition to \textit{Time} - to graph-structured explanations, independently evaluating both edge-based and feature-based attributions to capture the dual nature of graph explanations; (ii) {\bf an exhaustive, multi-task benchmarking study}, in which we evaluate 8 strictly vetted explainers across 10 GraphML tasks and multiple GNN architectures, generating ${\sim}2$M explanations and revealing critical insights such as the generally high stability but severely lacking pertinence of various widely adopted methods; and (iii) {\bf operational guidelines} where we provide empirically derived criteria for metric acceptability to assist practitioners in explainer selection, alongside a structural approach to translating complex graph attributions into natural language for regulatory compliance.

\section{Related Work}
\label{sec:relw}
\spara{Graph Machine Learning (Graph ML)} utilizes GNNs to process non-Euclidean relational data via message-passing mechanisms~\cite{nandan2025graphxai}. Formally, a graph is defined as $G=(V,E)$, and its inputs are characterized by their dual nature: a feature matrix $X\in\mathbb{R}^{n\times m}$ representing the attributes of $n=|V|$ nodes with $m$ features, and an adjacency matrix $A\in\{0,1\}^{n\times n}$ capturing the topological structure. A GNN learns a mapping from this joint input space to a target prediction $y$, effectively modeling the function $(X, A) \rightarrow y$. The foundational formulation of GNNs was initially introduced to extend neural network models to process cyclic, directed, and undirected graph structures directly~\cite{scarselli2008graph}. Standard modern architectures include Graph Convolutional Networks (GCNs)~\cite{kipf2016semi}, Graph Attention Networks (GATs)~\cite{velickovic2017graph}, GraphSAGE~\cite{hamilton2017inductive}, Graph Isomorphism Networks (GINs)~\cite{xu2018powerful} and Spatial GCNs~\cite{danel2020spatial}. Recent advancements integrate self-supervised pre-training to establish Graph Foundation Models (GFMs)~\cite{zhao2025gfm}.

These architectures are deployed across various levels of granularity, typically categorized into node-level, edge-level, and graph-level tasks. At the node level, the primary objective is \textit{node classification}, where the model predicts the target label $y_v$ of a specific node $v \in V$ based on its feature vector $x_v$ and the broader structural context provided by the adjacency matrix $A$. At the link level, edge-classification models aim to infer the existence of an unobserved connection between a pair of nodes $u$ and $v$, a fundamental task for recommendation systems and knowledge graph completion. Finally, \textit{graph classification} operates at the macroscopic level, predicting a global property $y_G$ for the entire graph. This requires the application of a readout or pooling function to aggregate localized node embeddings into a comprehensive global representation. All three tasks have their regression counterpart when the target $y$ is continuous.

\spara{G-XAI} considers explainers that identify the topological substructures, nodes, and features determining GNN predictions~\cite{nandan2025graphxai}. These methods are categorized into post-hoc and self-interpretable models that generate factual or counterfactual explanations. Post-hoc factual methods are primarily driven by attribution or perturbation mechanisms, often adapting general deep learning techniques to the graph domain. Gradient-based techniques leverage the model internal gradients to assign importance scores; these include fundamental approaches such as Saliency maps~\cite{simonyan2013deep}, Input $\times$ Gradient~\cite{shrikumar2017learning}, and Integrated Gradients~\cite{sundararajan2017axiomatic}, which aggregate gradients along a path from a baseline to the input. 
Advanced backpropagation variants like Guided Backpropagation~\cite{springenberg2014striving} and Deconvolution~\cite{zeiler2014visualizing} highlight influential features by suppressing negative gradients, while Layer-wise Relevance Propagation (LRP)~\cite{bach2015pixel} has been specifically reformulated for graphs as GNN-LRP~\cite{schnake2021higher} to decompose the final prediction score into relevance contributions that flow through the message-passing layers. In contrast, perturbation-based algorithms identify influential subgraphs by observing changes in the output when the input structure is masked.
GNNExplainer~\cite{ying2019gnnexplainer} learns a soft mask over the adjacency matrix and the node features to identify an influential subgraph, while PGExplainer~\cite{luo2020parameterized} extends this through a parameterized approach to produce global explanations. GraphMask~\cite{schlichtkrull2020interpreting} introduces a differentiable gating mechanism to selectively erase unnecessary edges during the message-passing process and revealing which path are actually essential for the model's prediction. Other structural methods include GraphSHAP~\cite{perotti2023graphshap}, which adapts Shapley values to graph neighborhoods, and SubgraphX~\cite{yuan2021subgraphx}, which uses Monte Carlo tree search to identify high‑value subgraphs under a Shapley‑based objective. Counterfactual explainers such as CF‑GNNExplainer~\cite{lucic2022cf} and CLEAR~\cite{ma2022clear} characterize model reasoning by identifying minimal topological perturbations that alter the prediction. Conversely, self-interpretable models constrain representations during training to produce inherently interpretable latent spaces, offering explanations via prototypes~\cite{zhang2022protgnn} or high-level concepts~\cite{azzolin2023global}.

%alan: qui vedere se asciugare un po'
\spara{G-XAI Evaluation} requires standardized metrics and frameworks. Key metrics include Fidelity, Sparsity, Stability, Graph Explanation Accuracy (GEA), Graph Explanation Faithfulness (GEF), and Graph Explanation Fairness (GECF) \cite{agarwal2023graphxai}. Benchmarking frameworks systematize these evaluations but differ markedly in scope and assumptions. GraphXAI uses the ShapeGGen generator to synthesize graphs with planted ground-truth explanations~\cite{agarwal2023graphxai}, while BAGEL's plausibility and correctness metrics likewise require reference rationales~\cite{rathee2022bagel}; both presuppose a known ground truth. GraphFramEx instead evaluates post-hoc methods on real-world graphs via a fidelity-based Characterization Score~\cite{amara2022graphframex}, and GnnX-Bench studies the stability and architectural sensitivity of perturbation-based explainers~\cite{kosan2024gnnxbench}. More specialized efforts target molecular datasets (B-XAIC~\cite{proszewska2025b}) and medical reasoning models (CURE-Bench~\cite{gao2025curebench}).
Our framework differs along three axes. First, it is \emph{ground-truth-free}: instead of scoring agreement with a planted rationale (as in GraphXAI and BAGEL), our metrics measure an explanation's consistency with the model's own behaviour, applying even to proprietary data with no reference explanation. Second, it spans a broader \emph{task spectrum}: prior benchmarks target classification, mostly at the node level, whereas we also cover regression and edge-level tasks, scoring feature and edge attributions jointly. Third, it targets \emph{production constraints}: we select explainers by software license and dual-attribution support, scale to ${\sim}2$M explanations via statistical early stopping, and distill the results into operational thresholds --- aspects often absent from existing frameworks.

\section{Metrics for G-XAI Evaluation}
\label{sec:metrics}

In this section, we describe our G-XAI evaluation framework. We define metrics to cover both node features and graph structure, and introduce optimizations to make them computationally tractable at industrial scale.
Following the approach in \cite{perotti2024explainability}, we identify four main qualities an explanation should possess for industrial applications: (i) algorithmic reproducibility and quasi-determinism; (ii) low cognitive load for a human; (iii) adherence to the model's inner mechanism; (iv) low execution time to produce the explanation.
The four quality dimensions we consider follow \cite{perotti2024explainability}, but each
operationalises a long-established explanation-quality property rather than a new one.
Effective Compactness is adapted from the tabular metric of \cite{perotti2024explainability};
Stability reuses its same-input formulation of the classic consistency/robustness criterion;
Pertinence is a faithfulness measure; and Time is a standard efficiency measure. Our
contribution is to adapt these properties to the dual nature of graph data - scoring edge
and feature attributions independently - and to keep them ground-truth-free.
Each quality dimension is evaluated through a dedicated metric, described subsequently. Every metric has two implementations depending on whether it operates on the node feature attribution matrix ${f}\in\mathbb{R}^{n\times m}$ (with $n$ the number of nodes, and $m$ the number of features) or the edge attribution $s\in\mathbb{R}^{|E|}$, with slight variations adopted per task. Execution time is measured directly during explanation generation. Time is a standard efficiency measure: the wall-clock runtime required to produce the
explanation. Pseudocode is provided for each metric.

\spara{Stability}
captures how deterministic an explainer is: given the same datapoint and model, does the explainer produce consistent explanations? We adapt this same-input formulation of Stability from the tabular benchmark of
\cite{perotti2024explainability}. It parallels Graph Explanation Stability~\cite{agarwal2023graphxai}
and, unlike input-perturbation robustness, serves to isolate the non-deterministic
explainers that must be invoked several times. We measure this by generating multiple pairs of explanations for the same input, computing their similarity, and averaging across pairs - as formalised in Algorithm~\ref{alg:stability_attribution}.
Let $G = (V, E)$ represent a graph, $M$ a model, and $t$ an explanation target. We generate $N$ independent pairs of explanations for the input $(t, G, M)$. 
We denote the sequence of explanation pairs as $((f_1^{(k)}, s_1^{(k)}), (f_2^{(k)}, s_2^{(k)}))_{k=1}^N$, with $f$ the feature attribution matrix and $s$ the edge attribution vector.
%\FloatBarrier
%\vspace*{2mm}
% Stability Algorithm
\begin{algorithm}[t!]
\small
\caption{Stability}
\label{alg:stability_attribution}
\SetKwInOut{Input}{Input}
\SetKwInOut{Output}{Output}

\Input{target $t$, graph $G$, model $M$, Explainer $\mathcal{E}$, task\_type, EarlyStop}
\Output{feat\_stability, edge\_stability}

$\mathcal{L}_{feat}, \mathcal{L}_{edge} \gets [ ],\, [ ]$ \tcp*{Empty lists}
$i \gets 0$\;

\While{$\neg (\text{EarlyStop}(i, \mathcal{L}_{feat}) \land \text{EarlyStop}(i, \mathcal{L}_{edge}))$}{
    $({f}_j, s_j) \gets \mathcal{E}(t, G, M) \quad \forall\, j \in \{1,2\}$ \tcp*{Two attribution passes}
    $\mathcal{L}_{feat}.\text{Append}\!\bigl(\text{StabilityPairFeat}({f}_1, {f}_2, t, M, \text{task\_type}, G)\bigr)$ \tcp*{Alg.~2}
    $\mathcal{L}_{edge}.\text{Append}\!\bigl(\text{StabilityPairEdge}(s_1, s_2, t, M, \text{task\_type}, G)\bigr)$ \tcp*{Alg.~3}
    $i \gets i + 1$\;
}

\Return{$\text{Average}(\mathcal{L}_{feat}),\; \text{Average}(\mathcal{L}_{edge})$}\;
\end{algorithm}

% %\FloatBarrier
% %\vspace*{5mm}
In edge attributions (Algorithm~\ref{alg:stability_pair_edge}), for a pair $s_1$ and $s_2$, we restrict the vectors to $E_{rel}$ (the edges in the computational graph of $t$), and apply the L2 normalization as $\hat{s}_i = \frac{s_i[E_{rel}]}{\|s_i[E_{rel}]\|_2}$ for $i \in \{1, 2\}$. For each pair we compute
$\rho_{edge}(s_1, s_2) = 1 - \frac{\|\hat{s}_1 - \hat{s}_2\|_2}{2}$, and the overall stability is the average of these pairwise similarities:
$$\text{Stability}_{edge}(\mathcal{E}, t, M, G) = \frac{1}{N} \sum_{k=1}^N \rho_{edge}(s_1^{(k)}, s_2^{(k)})$$

For feature attributions (Algorithm~\ref{alg:stability_pair_feature}), we partition each $f_i$  into $f_{A,i}$, which comprises the attributions of the target $t$, and $f_{B,i}$, which encompasses all remaining nodes within the computational graph. Both components are normalized (L2 element-wise norm) into $\hat{f}_{A,i}$ and $\hat{f}_{B,i}$, and we then compute $d_A$ and $d_B$ as: $d_X = \frac{\|\hat{f}_{X,1} - \hat{f}_{X,2}\|_2}{2} \quad \text{for } X \in \{A, B\}$. For each pair we then define: $\rho_{feat}(f_1, f_2) = 1 - \frac{d_A + d_B}{2}$, and we compute their average to obtain:
$$\text{Stability}_{feat}(\mathcal{E}, t, M, G) = \frac{1}{N} \sum_{k=1}^N \rho_{feat}(f_1^{(k)}, f_2^{(k)})$$

%\vspace*{-8mm}
% Stability Pair Edge Algorithm
\begin{algorithm}[t!]
\small
\caption{StabilityPairEdge}
\label{alg:stability_pair_edge}
\SetKwInOut{Input}{Input}
\SetKwInOut{Output}{Output}

\Input{$s_1$, $s_2$, target $t$, model $M$, task\_type, graph $G$}
\Output{edge\_stability\_pair}

\BlankLine
\uIf{task\_type == ``node''}{
    $E_{rel} \gets \text{GetEdgesInCompGraph}(t, M, G)$\;
}
\uElseIf{task\_type == ``link''}{
    $(n_1, n_2) \gets t$\;
    $E_{rel} \gets \text{GetEdgesInCompGraph}(n_1, M, G) \cup \text{GetEdgesInCompGraph}(n_2, M, G)$\;
}

$a_i \gets s_i[E_{rel}] / \|s_i[E_{rel}]\|_2 \quad \forall\, i \in \{1,2\}$ \tcp*{Filter \& Norm.}

\Return{$1 - \|a_1 - a_2\|_2 / 2$}\;
\end{algorithm}

%\vspace*{-12mm}

% Stability Pair Feature Algorithm
\begin{algorithm}[t!]
\small
\caption{StabilityPairFeature}
\label{alg:stability_pair_feature}
\SetKwInOut{Input}{Input}
\SetKwInOut{Output}{Output}

\Input{${f}_1$, ${f}_2$, target $t$, model $M$, task\_type, graph $G$}
\Output{feat\_stability\_pair}

\BlankLine
\uIf{task\_type == ``node''}{
    $V_{neigh} \gets \text{GetNodesInCompGraph}(t, M, G)$\;
    $f_{A,i} \gets {f}_i[t] \quad \forall\, i \in \{1,2\}$\;
}
\uElseIf{task\_type == ``link''}{
    $(v_1, v_2) \gets t$\;
    $V_{neigh} \gets \text{GetNodesInCompGraph}(v_1, M, G) \cup \text{GetNodesInCompGraph}(v_2, M, G)$\;
    $f_{A,i} \gets {f}_i[v_1] \oplus {f}_i[v_2] \quad \forall\, i \in \{1,2\}$\;
}

$f_{B,i} \gets \text{Flatten}({f}_i[V_{neigh}]) \quad \forall\, i \in \{1,2\}$\;
$f_{X,i} \gets f_{X,i} / \|f_{X,i}\|_2 \quad \forall\, X \in \{A,B\},\; i \in \{1,2\}$ \tcp*{L2 Norm.}
$d_X \gets \|v_{X,1} - v_{X,2}\|_2 / 2 \quad \forall\, X \in \{A,B\}$\;

\Return{$1 - (d_A + d_B) / 2$}\;
\end{algorithm}

%\vspace*{-5mm}

\spara{Effective Compactness}
(EC) captures the amount of information in the explanation required to impact the black-box model significantly. 
EC is adapted from \cite{perotti2024explainability} and operationalises the classic
conciseness/Sparsity property~\cite{amara2022graphframex}; it is conceptually related to
the Effective Complexity of Nguyen and Martinez~\cite{nguyen2020quantitative}, but defined
here in counterfactual terms, as the number of modifications needed to flip the prediction.
However, in graphs, an explanation can easily involve thousands of different node-feature pairs and edges: since EC is mostly relevant for human interactions with the model, excessively large values of the metric can be considered equivalent. For a target $t$ with computational graph $G_{target}$ and prediction as $\hat{y} = M(t, G_{target})$, EC computes the number of iterative modifications, bounded by a maximum of 100, needed to change $\hat{y}$ (i.e., the class is modified, or the regression output surpasses a threshold).

On edge attributions (Algorithm~\ref{alg:ec_edge_attribution}), we first sort the edges in the computational graph $E_{rel}$ by descending attribution score, obtaining the sequence $I_{sorted}$. 
%We iteratively remove edges from $G_{target}$ according to $I_{sorted}$. 
Let $G_{mod}^{(i)}$ represent the graph obtained after removing the first $i$ edges of $I_{sorted}$ from $G_{target}$. The edge EC is the smallest number of removals required to alter the original prediction, bounded by the maximum allowed iterations and the total size of $E_{rel}$:
$$\text{EC}_{edge}(s, t, M, G) = \min \left( \{ i \ge 1 \mid M(t, G_{mod}^{(i)}) \neq \hat{y} \} \cup \{ 100, |E_{rel}| \} \right)$$

For feature EC (Algorithm~\ref{alg:ec_feature_attribution}), modifications are performed via substitutions from reference graphs. We identify a set of reference computational graphs $\mathcal{G}_{ref}$ from medoid targets $t_{ref}$ whose baseline predictions $\hat{y}'$ differ from $\hat{y}$. In a regression task, we select medoids from the sets of nodes or edges that fall one standard deviation above and below the target prediction, while in a classification task we extract one medoid for class. Let $V_{neigh}$ be the nodes in $G_{target}$ different from $t$.
We split $f$ in the target ($f_{target}$) and neighbor ($f_{neighs}$) attribution vectors. For a node $t$ we have $f_{target}=f[t]\in\mathbb{R}^m$; for edge targets $t=(v_1, v_2)$, we define it by concatenation: $f_{target}=f[v_1] \oplus f[v_2]\in\mathbb{R}^{2m}$.
We obtain $f_{neighs}$ by performing a max pooling across nodes in $V_{neigh}$. We then construct a priority vector by concatenation ($f_C = f_{target} \oplus f_{neighs}$) and compute $I_{sorted}$, the indices which sort $f_C$ in descending order. For each reference target $t_{ref}$ and corresponding $G_{ref}$, we establish a random mapping $\phi$ from $V_{neigh}$ to the nodes in the neighborhood of $t_{ref}$. Then, we perform iterative substitutions according to $I_{sorted}$. When the $i$-th element of $I_{sorted}$  corresponds to a target feature, it is replaced with the corresponding value from $t_{ref}$. If instead it corresponds to a neighbor feature, the substitution is applied across all nodes in $V_{neigh}$ using the mapping $\phi$. If $G_{mod}^{(i)}(G_{ref})$ denote the modified graph after $i$ substitutions using $G_{ref}$, we define $k(G_{ref}) = \min \left( \{ i \ge 1 \mid M(t, G_{mod}^{(i)}(G_{ref})) \neq \hat{y} \} \cup \{ 100, |f_C| \} \right)$. The feature EC is defined as the minimum across all $G_{ref}$:
$$\text{EC}_{feat}(f, t, M, G) = \min_{G_{ref} \in \mathcal{G}_{ref}} k(G_{ref})$$

%\vspace*{-8mm}
\FloatBarrier
\begin{algorithm}[t!]
\small
\caption{ECEdge}
\label{alg:ec_edge_attribution}
\SetKwInOut{Input}{Input}
\SetKwInOut{Output}{Output}

\Input{Edge attribution $s$, target $t$, graph $G$, model $M$}
\Output{edge\_ec}

$E_{rel} \gets \text{GetEdgesInCompGraph}(t, M, G)$\;
$I_{sorted} \gets \text{ArgSort}(s[E_{rel}])$ \tcp*{Sort edges by attribution}
$G_{target} \gets \text{GetSubgraph}(G, t, M)$\;
$G_{mod} \gets G_{target}$\;
$i \gets 0$\;

\While{$M(t, G_{mod}) = M(t, G_{target}) \land i < \min(100, |E_{rel}|)$}{
    $G_{mod} \gets \text{RemoveEdge}(I_{sorted}[i], G_{mod})$\;
    $i \gets i + 1$\;
}

\Return{$i$}\;
\end{algorithm}

%\vspace*{5mm}

\begin{algorithm}[t!]
\small
\caption{ECFeature}
\label{alg:ec_feature_attribution}
\SetKwInOut{Input}{Input}
\SetKwInOut{Output}{Output}

\Input{Attribution matrix ${f}$, target $t$, graph $G$, model $M$, \# features $N_f$}
\Output{feature\_ec}

$V_{neigh} \gets \text{GetNodesInCompGraph}(t, M, G)$\;
$f_{C} \gets {f}[t] \oplus \text{Max}({f}[V_{neigh}], \text{dim}=1)$ \tcp*{${f}[v_1]\!\oplus\!{f}[v_2]$ for links}
$I_{sorted} \gets \text{ArgSort}(f_{C})$\;
$\mathcal{G}_{ref} \gets \text{GetRefGraphs}(G)$\;
$G_{target} \gets \text{GetSubgraph}(G, t, M)$\;
$\text{feature\_ec} \gets 100$\;

\ForEach{$(t_{ref}, G_{ref}) \in \mathcal{G}_{ref}$ \textbf{s.t.} $M(t_{ref}, G_{ref}) \neq M(t, G_{target})$}{
    $\phi \gets \text{CreateNodeMapping}(G_{target}, G_{ref})$\;
    $G_{mod} \gets G_{target}$\;
    $i \gets 0$\;
    \While{$M(t, G_{mod}) = M(t, G_{target}) \land i < \min(100, |f|)$}{
        \eIf{$I_{sorted}[i] < |f|-N_f$}{
            $G_{mod} \gets \text{SetTargetFeat}(I_{sorted}[i], G_{mod}, G_{ref}, \phi)$\;
        }{
            $G_{mod} \gets \text{SetNeighsFeats}(I_{sorted}[i] \bmod N_f, G_{mod}, G_{ref}, \phi)$\;
        }
        $i \gets i + 1$\;
    }
    $\text{feature\_ec} \gets \min(\text{feature\_ec}, i)$\;
}
\Return{feature\_ec}\;
\end{algorithm}

%\FloatBarrier
%\vspace*{5mm}

\spara{Pertinence} addresses the adherence of the explanation to the actual model. 
Pertinence is a faithfulness measure in the family of Fidelity and Graph Explanation
Faithfulness (GEF)~\cite{agarwal2023graphxai}. It adapts the rank-based idea of the Rank
Quality Index of \cite{perotti2024explainability} and, like GraphFramEx~\cite{amara2022graphframex},
is ground-truth-free, but it additionally scores the attribution ordering against random
baselines.
A highly-pertinent explanation should impact the black-box model more than any randomly-generated one, while a low-pertinence one is indistinguishable from (or worse than) random baselines. 
We use the iterative modification procedures established for EC. Let $K$ be the EC score ($\text{EC}_{edge}$ or $\text{EC}_{feat}$). We define a deletion curve $C(k)$ as the model's output after $k$ successive modifications ($1 \le k \le K$). We define $C_{del}$ as the deletion curve constructed according to the explanation ranking $I_{sorted}$, while $\mathcal{C}_{rand} = \{C_{rand}^{(j)}\}_{j=1}^N$ is a set of $N$ random curves, generated by applying modifications according to a random permutation of $I_{sorted}$. The pertinence score relative to a single $C_{rand}^{(j)}$ is the proportion of points where $C_{del}$ degrades the output \emph{strictly faster}: $p(C_{del}, C_{rand}) = \frac{1}{K} \sum_{k=1}^K \mathbb{I}(C_{del}(k) < C_{rand}(k))$.

For edge attributions (Algorithm~\ref{alg:pertinence_edge}), the modifications are sequential edge removals from $G_{target}$. The edge pertinence is the average of $p(C_{del}, C_{rand})$:
$$\text{Pertinence}_{edge}(s, t, M, G) = \frac{1}{N} \sum_{j=1}^N p(C_{del}, C_{rand}^{(j)})$$

For feature attributions (Algorithm~\ref{alg:pertinence_feature}), modifications consist of feature substitutions. We construct a deletion curve $C_{del}(G_{ref})$ and a set of random curves $\mathcal{C}_{rand}(G_{ref})$ for each reference graph $G_{ref} \in \mathcal{G}_{ref}$. The feature pertinence is the average success ratio across all random permutations and reference graphs:$$\text{Pertinence}_{feat}(f, t, M, G) = \frac{1}{|\mathcal{G}_{ref}| N} \sum_{G_{ref} \in \mathcal{G}_{ref}} \sum_{j=1}^N p(C_{del}(G_{ref}), C_{rand}^{(j)}(G_{ref}))$$

%\FloatBarrier
%\vspace*{5mm}
\begin{algorithm}[t!]
\small
\caption{PertinenceEdge}
\label{alg:pertinence_edge}
\SetKwInOut{Input}{Input}
\SetKwInOut{Output}{Output}

\Input{Edge attribution $s$, edge\_ec, target $t$, graph $G$, model $M$}
\Output{edge\_pertinence}

$E_{rel} \gets \text{GetEdgesInCompGraph}(t, M, G)$\;
$I_{sorted} \gets \text{ArgSort}(s[E_{rel}])$\;
$C_{del} \gets \text{CreateEdgeDelCurve}(I_{sorted}, \text{edge\_ec}, t, G, M)$ \tcp*{Deletion curve}
$\mathcal{L}_{pert} \gets [ ]$\;
$i \gets 0$\;

\While{$\neg \text{EarlyStop}(i, \mathcal{L}_{pert})$}{
    $C_{rand} \gets \text{CreateEdgeDelCurve}(\text{RandPerm}(I_{sorted}), \text{edge\_ec}, t, G, M)$\;
    $\mathcal{L}_{pert}.\text{Append}(\text{CompareCurves}(C_{del}, C_{rand}))$\;
    $i \gets i + 1$\;
}

\Return{$\text{Average}(\mathcal{L}_{pert})$}\;
\end{algorithm}

\begin{algorithm}[t!]
\small
\caption{PertinenceFeature}
\label{alg:pertinence_feature}
\SetKwInOut{Input}{Input}
\SetKwInOut{Output}{Output}

\Input{Attribution matrix ${f}$, feat\_ec, target $t$, graph $G$, model $M$}
\Output{feature\_pertinence}

$V_{neigh} \gets \text{GetNodesInCompGraph}(t, M, G)$\;
$I_{sorted} \gets \text{ArgSort}({f}[t] \oplus \text{Max}({f}[V_{neigh}], \text{dim}\!=\!1))$\;
$\mathcal{G}_{ref} \gets \text{GetRefGraph}(G, \text{task\_type})$\;
$\mathcal{L}_{pert} \gets [ ]$\;

\ForEach{$(t_{ref}, G_{ref}) \in \mathcal{G}_{ref}$ \textbf{s.t.} $M(t_{ref}, G_{ref}) \neq M(t, G_{target})$}{
    $C_{del} \gets \text{CreateFeatDelCurve}(I_{sorted}, \text{feat\_ec}, t, G_{ref}, G, M)$\;
    $\mathcal{L}_{curr} \gets [ ]$\;
    $i \gets 0$\;
    \While{$\neg \text{EarlyStop}(i, \mathcal{L}_{curr})$}{
        $C_{rand} \gets \text{CreateFeatDelCurve}(\text{RandPerm}(I_{sorted}), \text{feat\_ec}, t, G_{ref}, G, M)$\;
        $\mathcal{L}_{curr}.\text{Append}(\text{CompareCurves}(C_{del}, C_{rand}))$\;
        $i \gets i + 1$\;
    }
    $\mathcal{L}_{pert}.\text{Append}(\text{Average}(\mathcal{L}_{curr}))$\;
}

\Return{$\text{Average}(\mathcal{L}_{pert})$}\;
\end{algorithm}

%\FloatBarrier
%\vspace*{5mm}

\spara{Computational Optimisation}
While our metrics do not introduce computational overhead beyond the explainer itself, repeated runs for stability and pertinence estimation can be costly in time-sensitive pipelines. To mitigate this, we implement an early stopping strategy that monitors the convergence of the observed distribution across runs, terminating evaluation once the estimate has stabilized - reducing wall-clock time without sacrificing accuracy.

Let $\mathcal{M} = \{m_1, m_2, \dots, m_i\}$ 
be the set of stability/pertinence values collected after $i$ independent trials. For 
$i \ge 30$, we model the confidence interval of the mean $\mu_{\mathcal{M}}$ 
using the standard error $SE = \sigma_{\mathcal{M}} / \sqrt{i}$,
where $\sigma_{\mathcal{M}}$ represents the empirical standard deviation. 
We define two convergence tests based on a 95\% confidence level:
\begin{enumerate}
    \item \textbf{Mean Convergence:} Given a threshold $\tau$, we perform a 
    one-tailed z-test where the process terminates if 
    $z = \frac{|\mu_{\mathcal{M}} - \tau|}{SE} > 1.645$.
    \item \textbf{Precision Tolerance:} To ensure the stability of the 
    estimate, we stop if the standard error falls within a predefined 
    tolerance $\epsilon$: $z = \frac{\epsilon}{SE} > 1.96$, corresponding 
    to a two-tailed test.
\end{enumerate}
In all experiments we set $\tau = 0.5$ (the random-baseline expectation for   
  Pertinence) and $\epsilon = 0.05$. We begin checking convergence after        
  $i = 30$ trials and cap at $N = 100$.
In our benchmarking, this approach reduced the required iterations from a 
fixed $N=100$ to an average of 35-40, yielding a significant reduction in 
wall-clock time with minimal impact on estimation variance.
The metrics differ in evaluation cost. Stability re-invokes the explainer N times, so its cost scales with the explainer's runtime (negligible for gradient methods, high for GNNExplainer and GraphMask). Effective Compactness is cheap, requiring a single deletion sequence of at most 100 model inferences, whereas Pertinence is the most expensive: it repeats that sequence over N random baselines, at a cost on the order of N times Effective Compactness. The early-stopping scheme above targets precisely the repeated estimation in Stability and Pertinence.

\section{G-XAI Library}
\begin{figure}[t!]
    \centering
    \includegraphics[width=\textwidth]{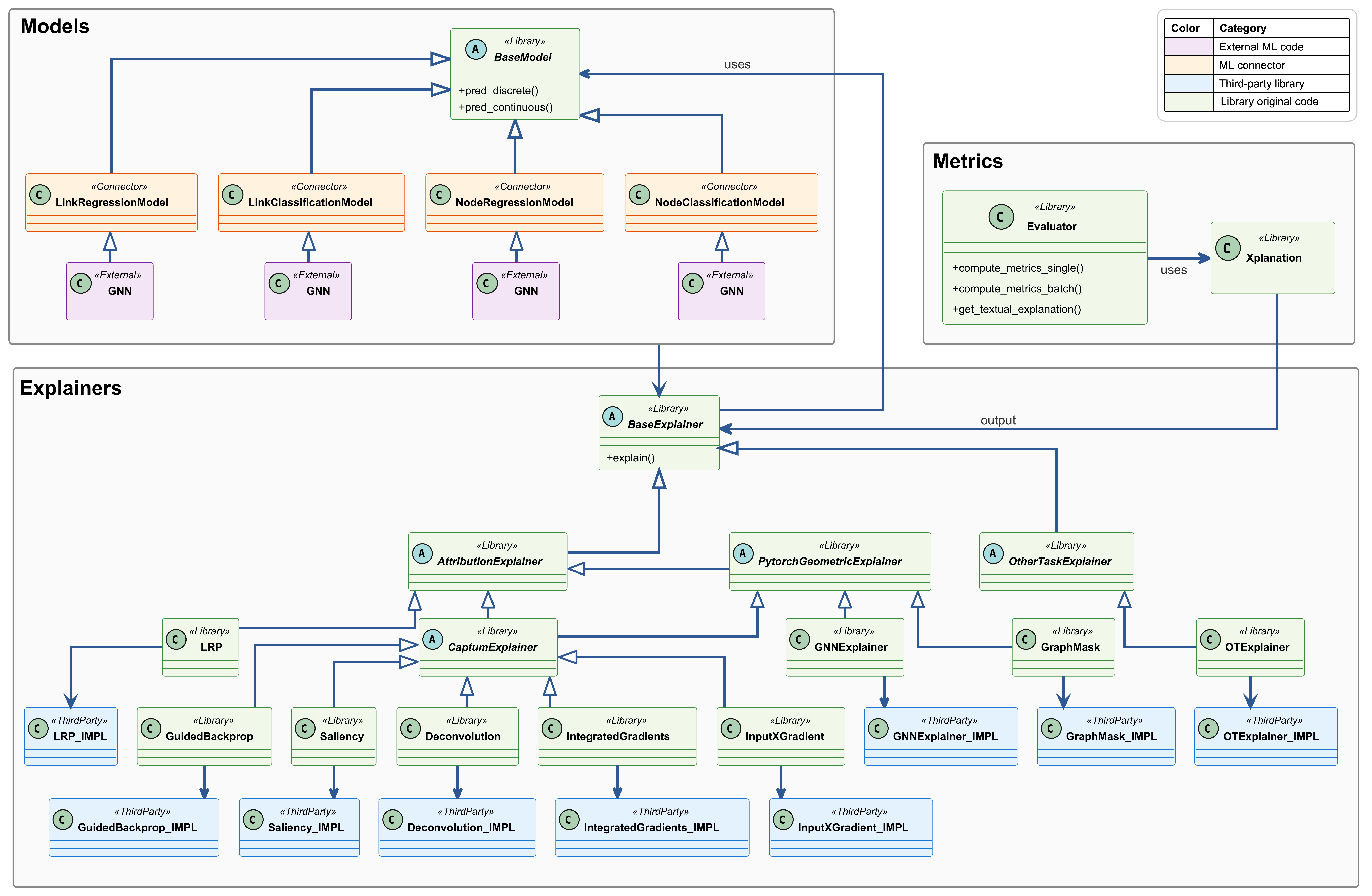}
    \caption{Architectural overview of the proposed XAI framework, illustrating the decoupling between core evaluation logic (green), ML model connectors (orange), and the explainer suite (purple models and blue explainers).}
    \label{fig:uml_diagram} 
    \vspace*{-3mm}
\end{figure}

The proposed framework is implemented as a modular Python library designed to make GNN explanation quality measurable within existing ML pipelines with minimal integration overhead. The library is proprietary and cannot be released as open source due to company copyright restrictions. To support reproducibility and follow-up research within these constraints, the authors can be contacted for access to selected snippets of code. Figure \ref{fig:uml_diagram} illustrates the overall architecture, which is built upon the PyTorch Geometric module and follows an object-oriented design to ensure modularity and extensibility.

\spara{Core Abstractions}
The library's design is centered around four fundamental abstraction layers:
\begin{itemize}
    \item \texttt{BaseModel}: An abstract wrapper that standardizes the 
    interface for diverse GNN tasks. It handles the dual nature of graph 
    inputs (feature matrix $X \in \mathbb{R}^{n \times f}$ and adjacency 
    matrix $A \in \{0, 1\}^{n \times n}$) and provides uniform methods for 
    both discrete (\texttt{pred\_discrete}) and continuous 
    (\texttt{pred\_continuous}) outputs.
    \item \texttt{BaseExplainer}: A high-level interface for explanation 
    generation. We provide specialized subclasses for \textit{Attribution 
    Methods} (e.g., Integrated Gradients, LRP), while the architecture is 
    designed to accommodate additional explainer families via the 
    \texttt{OtherTaskExplainer} extension point.
    \item \texttt{Xplanation}: A container class that encapsulates generated 
    attribution matrices alongside metadata regarding the target node/link 
    and the explainer's hyperparameters. It also stores computed metrics to 
    avoid redundant recomputations.
    \item \texttt{Evaluator}: The orchestration layer responsible for metric 
    computation. It facilitates batch processing of metrics and integrates a 
    Natural Language eXplanation (NLX) module to translate technical importance 
    scores into human-intelligible summaries.
\end{itemize}

\spara{Integration and Performance}
The library implements custom model wrappers for \textit{Node} and \textit{Link} tasks across both classification and regression settings. 
For link-level tasks, the library automatically handles the generation of computational subgraphs for the target edge $(u, v)$, ensuring that the explanation captures the relational context of both endpoints. This is a non-trivial requirement in link prediction tasks where neither node alone determines the prediction. The library supports hardware acceleration (CUDA) and provides automated utilities for mapping neighborhood-level feature 
attributions back to a target-centric representation, making the output immediately interpretable within standard ML workflows.

\spara{Natural Language Explanations} 
To bridge the gap between technical attributions and regulatory compliance, the library includes a Natural Language eXplanation (NLX) module via the \texttt{Evaluator.get\_textual\_explanation()} method. Given an \texttt{Xplanation} object, the module: (i) ranks the top-$k$ influential features and edges by attribution magnitude; (ii) maps indices to human-readable names using a user-provided dictionary; and (iii) composes them into natural language via customizable templates. For instance, in a loan approval task, the module might output:
\textit{``The application was denied primarily due to a high Debt-to-Income Ratio (importance: 0.73). While Credit History was positive, the Debt level exceeded thresholds. The model also considered connections to 3 accounts with prior defaults.''}
\noindent
This approach ensures explanations remain faithful to the attributions while meeting EU AI Act transparency requirements. Templates support multi-lingual outputs and domain-specific terminology. We stress that the NLX module is a utility of the library for producing human-readable summaries; it is neither a benchmarked component nor part of the quantitative evaluation in Section 5.

\section{Experiments} % Frappa
\label{sec:setup}

As a first use of our framework, we perform a large-scale benchmark of explainers across a variety of tasks, datasets, and models. %In this section, we provide the experimental setup and the results. In the next section, instead, we discuss and analyse the metrics by performing a more in-depth comparison.

\subsection{Experimental setup}
For each task-dataset combination, we explain 100 target predictions (on either nodes or edges, according to the task), with the same targets across different models and explainers. We measure Pertinence, Effective Compactness, and Stability, alongside the time required by each explainer. We finally average the metrics across explanations, models, and datasets.

\spara{Datasets and Models} We selected five graph datasets for classification and regression, using each on both node-level and edge-level tasks - thus creating 10 GraphML tasks. For classification tasks we utilized Cora~\cite{yang2016revisiting}, GitHub~\cite{rozemberczki2019multi}, and BAShapes~\cite{ying2019gnnexplainer}. To address the scarcity of regression benchmarks, we curated a modified version of MovieLens~\cite{harper2015movielens} and generated a second fully synthetic dataset. For each dataset and task, we use two-layer GCNs and GAT. We provide additional details on both datasets and models in the Appendix.

\spara{Explainers} We initially identified many graph post-hoc explainers for our benchmark. However, in order to use them across different tasks and settings, we filtered them for their ability to handle both node and edge features, to explain classification and regression tasks on nodes and edges, and to output both feature and edge attributions. Due to the industrial application of our framework, we additionally had to exclude explainers without an explicit open-source software license. Many state-of-the-art graph explainers were excluded by these criteria, such as PGExplainer~\cite{luo2020parameterized}, SubgraphX~\cite{yuan2021subgraphx}, GNN-LRP~\cite{schnake2021higher}, GraphLIME~\cite{huang2022graphlime}, Degree~\cite{feng2022degree} and Clear~\cite{ma2022clear}. We ended in eight explainers for our benchmark: Saliency~\cite{simonyan2013deep}, Input$\times$Gradient~\cite{shrikumar2017learning}, Integrated Gradients~\cite{sundararajan2017axiomatic}, Guided Backpropagation~\cite{springenberg2014striving}, Deconvolution~\cite{zeiler2014visualizing}, Layer-wise Relevance Propagation (LRP)~\cite{bach2015pixel}, GNNExplainer~\cite{ying2019gnnexplainer}, and GraphMask~\cite{schlichtkrull2020interpreting}.
They fall into two categories: gradient-based (e.g. Saliency) or attribution-redistribution approaches (e.g. LRP), both of which compute importance by backpropagating signals from the output to the input, and perturbation-based masking approaches (e.g. GNNExplainer), that optimize "soft masks" to isolate the most influential subgraphs and features.

\subsection{Results}

\begin{table}[t!]
    \scriptsize
    \centering
    \caption{Benchmark results across models, datasets and tasks (best in bold).}
    \resizebox{\textwidth}{!}{
    \begin{tabularx}{1.3\textwidth}{l|lXXXXXXX}
    \toprule
     Task & Explainer & Feature \newline Pertinence & Edge \newline Pertinence & Feature \newline Eff. Comp. & Edge \newline Eff. Comp. & Feature \newline Stability & Edge \newline Stability & Execution \newline Time \\
    \midrule
    Node Class. & Deconv & 0.58±0.25 & 0.84±0.10 & 67.9±22.2 & 39.6±24.7 & \textbf{1.00±0.00} & \textbf{1.00±0.00} & \textbf{0.01±0.00} \\
    Node Class. & GNNExp & 0.90±0.10 & 0.54±0.12 & 32.2±12.8 & 42.8±23.0 & 0.97±0.01 & 0.96±0.03 & 0.87±0.10 \\
    Node Class. & GraphMask & 0.52±0.16 & 0.43±0.04 & 82.6±18.3 & 61.4±28.1 & 0.96±0.00 & 0.86±0.08 & 3.36±2.82 \\
    Node Class. & GuidedBP & 0.61±0.25 & 0.84±0.10 & 66.8±21.6 & 39.6±24.4 & \textbf{1.00±0.00} & \textbf{1.00±0.00} & \textbf{0.01±0.00} \\
    Node Class. & Input×Grad & \textbf{0.94±0.06} & 0.86±0.08 & \textbf{25.1±10.9} & 38.4±23.3 & \textbf{1.00±0.00} & \textbf{1.00±0.00} & \textbf{\textbf{0.01±0.00}} \\
    Node Class. & IntGrad & \textbf{0.94±0.06}& \textbf{0.87±0.08} & 25.5±11.2 & \textbf{37.0±22.3} & \textbf{1.00±0.00} & \textbf{1.00±0.00} & 0.32±0.09 \\
    Node Class. & LRP & 0.87±0.09 & 0.75±0.05 & 35.9±9.5 & 43.1±28.6 & \textbf{1.00±0.00} & \textbf{1.00±0.00} & \textbf{0.01±0.00} \\
    Node Class. & Saliency & 0.61±0.25 & 0.68±0.06 & 50.3±26.6 & 42.1±27.5 & \textbf{1.00±0.00} & \textbf{1.00±0.00} & \textbf{0.01±0.00} \\
    \midrule
    Node Reg. & Deconv & 0.63±0.09 & \textbf{\textbf{1.00±0.00}} & 21.9±36.6 & \textbf{24.5±32.5} & \textbf{\textbf{1.00±0.00}} & \textbf{\textbf{1.00±0.00}} & \textbf{0.02±0.01} \\
    Node Reg. & GNNExp & 0.67±0.16 & 0.48±0.12 & 30.9±43.0 & 40.7±38.5 & 0.96±0.00 & 0.98±0.00 & 2.44±0.73 \\
    Node Reg. & GraphMask & 0.53±0.05 & 0.48±0.03 & 34.7±56.1 & 56.3±37.8 & 0.97±0.00 & 0.88±0.14 & 7.89±5.61 \\
    Node Reg. & GuidedBP & 0.63±0.09 & \textbf{1.00±0.00} & 21.9±36.6 & \textbf{24.5±32.5} & \textbf{1.00±0.00} & \textbf{1.00±0.00} & 0.03±0.01 \\
    Node Reg. & Input×Grad & \textbf{0.90±0.07} & \textbf{1.00±0.00} & \textbf{7.0±9.9} & \textbf{24.5±32.5} & \textbf{1.00±0.00} & \textbf{1.00±0.00} & \textbf{0.02±0.01} \\
    Node Reg. & IntGrad & \textbf{0.90±0.07} & \textbf{1.00±0.00} & 7.3±11.0 & 24.9±33.2 & \textbf{1.00±0.00} & \textbf{1.00±0.00} & 0.93±0.34 \\
    Node Reg. & LRP & 0.75±0.17 & 0.74±0.25 & 20.5±20.7 & 37.0±30.1 & \textbf{1.00±0.00} & \textbf{1.00±0.00} & \textbf{0.02±0.02} \\
    Node Reg. & Saliency & 0.57±0.13 & 0.57±0.20 & 21.8±36.0 & 32.6±26.9 & \textbf{1.00±0.00} & \textbf{1.00±0.00} & 0.03±0.01 \\
    \midrule
    Edge Class. & Deconv & 0.56±0.25 & 0.81±0.25 & 86.9±14.9 & 50.7±14.9 & \textbf{1.00±0.00} & \textbf{1.00±0.00} & \textbf{0.02±0.01} \\
    Edge Class. & GNNExp & 0.76±0.26 & 0.54±0.21 & 44.7±19.3 & 63.3±18.6 & 0.97±0.00 & 0.96±0.02 & 1.75±0.86 \\
    Edge Class. & GraphMask & 0.31±0.24 & 0.31±0.21 & 90.8±5.2 & 82.9±13.7 & 0.96±0.00 & 0.80±0.15 & 12.43±18.19 \\
    Edge Class. & GuidedBP & 0.58±0.26 & 0.81±0.24 & 82.8±17.9 & 51.3±14.4 & \textbf{1.00±0.00} & \textbf{1.00±0.00} & \textbf{0.02±0.01} \\
    Edge Class. & Input×Grad & \textbf{0.87±0.20} & 0.83±0.25 & 28.7±21.0 & 50.9±15.0 & \textbf{1.00±0.00} & \textbf{1.00±0.00} & \textbf{0.02±0.01} \\
    Edge Class. & IntGrad & \textbf{0.87±0.20} & \textbf{0.84±0.25} & \textbf{27.7±21.8} & \textbf{48.7±15.0} & \textbf{1.00±0.00} & \textbf{1.00±0.00} & 0.73±0.34 \\
    Edge Class. & LRP & 0.60±0.25 & 0.58±0.18 & 73.4±18.5 & 63.9±12.6 & \textbf{1.00±0.00} & \textbf{1.00±0.00} & 0.03±0.01 \\
    Edge Class. & Saliency & 0.51±0.24 & 0.63±0.21 & 82.5±24.3 & 55.4±12.8 & \textbf{1.00±0.00} & \textbf{1.00±0.00} & \textbf{0.02±0.01} \\
    \midrule
    Edge Reg. & Deconv & 0.57±0.07 & 0.97±0.05 & 49.0±53.8 & \textbf{49.0±52.1} & \textbf{1.00±0.00} & \textbf{1.00±0.00} & \textbf{0.03±0.00} \\
    Edge Reg. & GNNExp & 0.60±0.13 & 0.60±0.22 & 51.2±55.2 & 74.6±35.9 & 0.96±0.01 & 0.99±0.01 & 2.74±0.44 \\
    Edge Reg. & GraphMask & 0.61±0.14 & 0.38±0.19 & 51.3±68.8 & 78.1±31.0 & 0.97±0.00 & 0.89±0.09 & 7.44±1.32 \\
    Edge Reg. & GuidedBP & 0.57±0.07 & 0.97±0.05 & 49.0±53.8 & \textbf{49.0±52.1} & \textbf{1.00±0.00} & \textbf{1.00±0.00} & \textbf{0.03±0.00} \\
    Edge Reg. & Input×Grad & \textbf{0.78±0.15} & 0.97±0.05 & \textbf{47.9±52.7} & \textbf{49.0±52.1} & \textbf{1.00±0.00} & \textbf{1.00±0.00} & \textbf{0.03±0.00} \\
    Edge Reg. & IntGrad & 0.77±0.15 & \textbf{0.98±0.03} & \textbf{47.9±52.7} & 49.1±52.2 & \textbf{1.00±0.00} & \textbf{1.00±0.00} & 1.32±0.02 \\
    Edge Reg. & LRP & 0.55±0.10 & 0.68±0.17 & 49.3±52.3 & 56.3±39.1 & \textbf{1.00±0.00} & \textbf{1.00±0.00} & \textbf{0.03±0.01} \\
    Edge Reg. & Saliency & 0.56±0.06 & 0.59±0.29 & 48.9±53.7 & 51.0±38.5 & \textbf{1.00±0.00} & \textbf{1.00±0.00} & \textbf{0.03±0.00} \\
    \bottomrule
    \end{tabularx}
    }
    \label{tab:benchmark}
\end{table}

\begin{figure}[t!]
    \centering
     \makebox[\textwidth][c]{\includegraphics[width=1\textwidth]{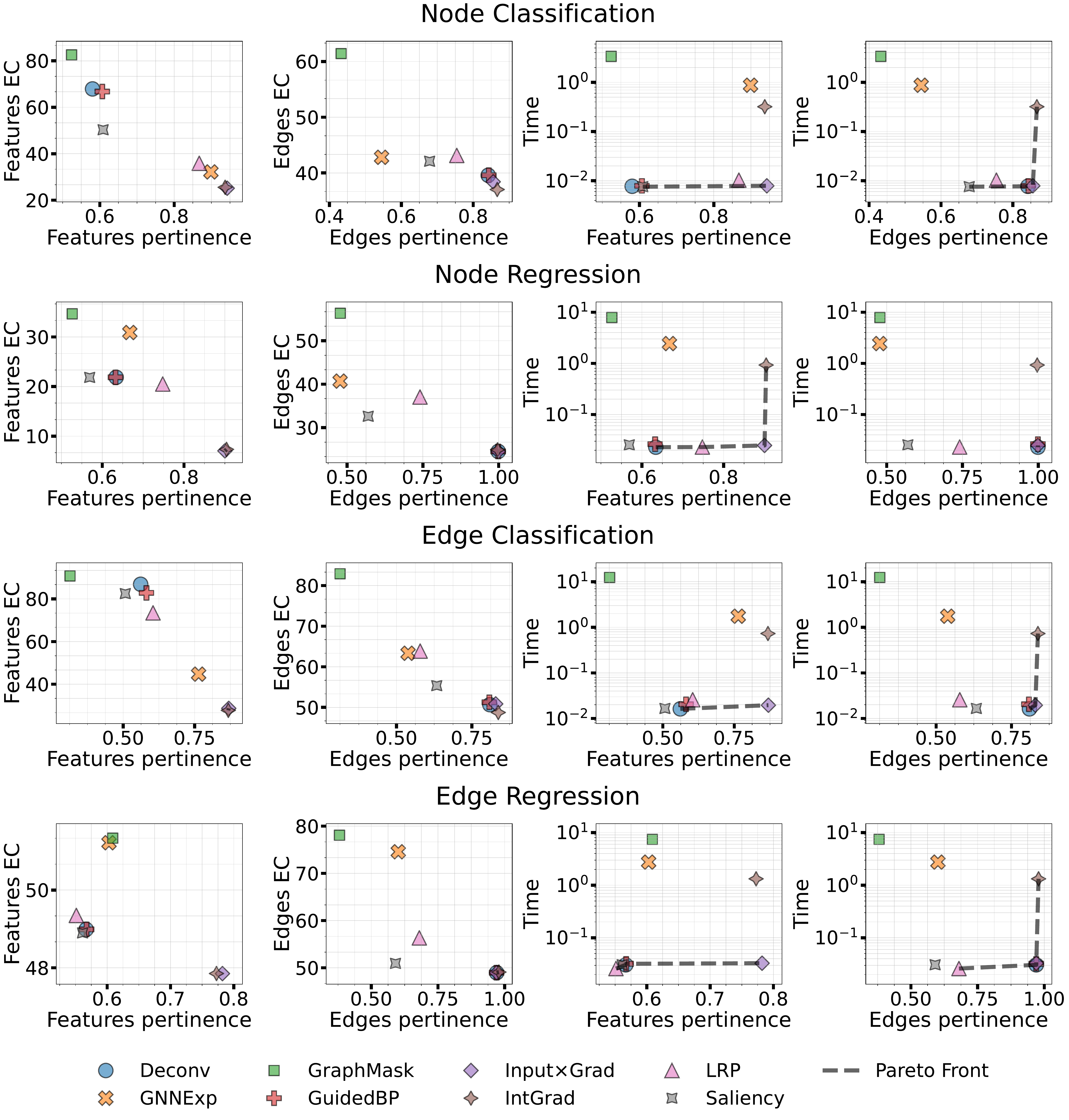}}

    \caption{Pareto plots comparing the explainers across pairs of metrics. In all cases, the bottom‑right region of the plot corresponds to more desirable trade‑offs.}
    \label{fig:pareto}
\vspace*{-5mm}
\end{figure}

We compare explainers across the four main tasks we considered. We average the metrics across explanation targets, datasets, and models, providing a robust empirical benchmark of the explainers' performances for each task. We report these results in Table~\ref{tab:benchmark}. Input$\times$Gradient and IntegratedGradients emerge as the top-performing explainers in effective compactness. In contrast, GraphMask consistently ranks as the least in the metric. The 100-iteration cap reflects a comprehensibility limit rather than a mere computational bound: an explanation that requires altering more than 100 elements to change the prediction is already beyond what a human can meaningfully inspect. Within this range, Effective Compactness rewards the virtuous explainers that reach the decision boundary by acting on the fewest elements.

The high performance of gradient methods in explanation fidelity/faithfulness to the model is also consistent with the G-XAI literature~\cite{amara2022graphframex,agarwal2023graphxai,proszewska2025b,nerini2025true,Monti2024ATA}.  Since pertinence evaluates how faithfully an explanation reflects model decisions, gradient methods benefit from direct access to internal weights, allowing for a fine-grained behavior analysis. In contrast, graph-based explainers like GNNExplainer and GraphMask do not have access to the internal model, and have to rely on random perturbations, introducing noise  in the explanations.

A subset of methods scores pertinence below the $0.5$ random baseline. This is not necessarily evidence of an attribution worse than random: coincident points of the explanation and random deletion curves are counted as failures, so tasks with few features (or many relevant ones) produce frequent ties that depress pertinence below $0.5$. A distinct, task-level effect arises when opposing classes are characterized by a multimodal distribution: the medoid reference targets then may poorly represent the target's nearest alternative and feature perturbations may lead to unstable prediction flips, letting random orderings degrade the output almost as fast as the attribution (we detail these cases in the Appendix). Computation variants that explicitly consider this phenomenon and %that ranks the explanation curve against the full population of random curves may raise the average pertinence and 
remove this sub-baseline regime are left to future work.

Notably, however, several methods exhibit specialized performance: GNNExplainer (in classification tasks) and LRP (in node tasks) excel at feature-level explanations, while Deconvolution and GuidedBackProp always perform better on edge attributions. In terms of stability, all explainers, with the exceptions of GNNExplainer and GraphMask, yield constant results due to their non-stochastic algorithms. A Stability of 1.00 means the explainer is fully deterministic, so a single invocation already yields a reproducible attribution. The metric's main practical value is thus to isolate the non-deterministic explainers (here GNNExplainer and GraphMask), which must be invoked several times before their attributions can be considered reliable.
Execution times, finally, vary drastically across the methods. Gradient-based methods are extremely efficient, typically requiring less than 0.01s, as they only need one model call. However, Integrated Gradients, GNNExplainer and GraphMask are significantly slower, lagging by two to four orders of magnitude, as they all require several model calls to generate an explanation.

The Pareto plots in Figure~\ref{fig:pareto} reveal no significant trade-off between pertinence and effective compactness: Input$\times$Gradient and Integrated Gradients function as dominant solutions across both metrics. This finding is consistent across most tasks, except for edge regression, where most methods show comparable performance regarding feature effective compactness. In this case, the feature pertinence provides a more nuanced evaluation, showing that Integrated Gradients and Input$\times$Gradient still perform significantly better than the random baseline (which, for the way we defined the metric, corresponds to $0.5$). We excluded stability due to its consistency across different explainers. 
When comparing execution time against feature and edge pertinence, instead, other gradient-based methods (LRP and Saliency) appear on the Pareto front due to their speed, with Input$\times$Gradient offering the best balance: it provides a substantial boost in pertinence for both edges and features with only a negligible increase in latency. Conversely, the marginal performance gains often offered by Integrated Gradients do not seem to justify an execution time that is two orders of magnitude higher. When considering edge pertinence, however, Deconvolution and Guided Backpropagation are almost on par with Input$\times$Gradient.
Finally, we observe that edges serve as the primary information source for models in regression tasks. However, this effect, alongside the limited variability in feature effective compactness in edge regression, may derive from the specific synthetic datasets used and might not generalize to other contexts.
%\FloatBarrier

\subsection{Operational Guidelines}

From our benchmarking results, we distill a set of practical guidelines for explainer selection and quality assessment, detailed in the Appendix. Amongst the most relevant components are empirically grounded quality thresholds, transferability considerations, and explainer recommendations.

\spara{Quality thresholds}
Based on our benchmarking results, we derive empirically grounded thresholds for Stability, Pertinence, and EC across all four Graph ML tasks, alongside transferability considerations and explainer recommendations. Stability thresholds transfer reliably across datasets, while Pertinence and Effective Compactness should be revalidated on proprietary data - particularly Edge EC, which varies with node degree. Full guidelines, including deployment workflows and tabular summaries, are provided in the Appendix.

\spara{Transferability notes.} Stability transfers reliably from benchmarks to new datasets, as it primarily depends on the explainer's algorithm. Pertinence and Effective Compactness are indicative but should be revalidated on proprietary data - particularly Edge EC, which varies with node degree. We recommend a minimum sample of 50--100 instances for initial assessment and 200+ for production validation.

\spara{Explainer recommendation.} For most graph tasks, \textbf{Input$\times$Gradient} offers the best trade-off: near-optimal pertinence and compactness with execution times under 0.03s. Integrated Gradients provides marginal gains at 100$\times$ higher latency, justified only when time is unconstrained. GraphMask consistently underperforms and should be avoided.

\section{Conclusion}
\label{sec:conclusions}

In this paper, we introduced an empirical evaluation framework for GNN post-hoc explanations designed for seamless application across diverse real-world datasets, models, and tasks. By implementing these metrics within a dedicated library, we enabled the integration and testing of ML models alongside a wide selection of explainers. Our benchmarking across both node and link tasks revealed that Input×Gradient (a relatively simple, deterministic, gradient-based technique) outperforms or matches the performance of more complex or graph-specific explainers. Our findings cross-validate results in the G-XAI benchmark literature and demonstrate a clear relationship between current fidelity metrics and our proposed Pertinence and Effective Compactness metrics.

The scope of our benchmark was constrained by our selection criteria: each explainer had to handle node and edge tasks under both classification and regression, and to produce feature and edge attributions simultaneously (dual-attribution). 
We further stress that our metrics deliberately measure consistency with the model's own behaviour rather than correctness with respect to real-world phenomena: in line with the true-to-the-model paradigm, it is the model, not an external oracle, that explanations must faithfully reflect. Validating explanations against domain knowledge is a complementary, human-centred direction that we leave to future work.
While necessary for a uniform comparison across tasks, these criteria introduce a selection bias, since several graph-native explainers do not meet them and thus fall outside our comparison. Nevertheless, our modular methodology provides a robust foundation for the future integration of new techniques and tasks. Subsequent research will expand this framework to include emerging graph explainability approaches, such as counterfactuals, while leveraging the insights from this benchmark to develop more flexible and effective graph explainers.

% Future work
% Limitations

%\bibliographystyle{splncs04} TODO: reinsert
\bibliographystyle{unsrt}
%% Note that this preceding line implies that you store your BibTeX references in a file called 'mybibliography.bib'. If you instead store your references in a file with a different name, for instance 'references.bib', the preceding line should read '\bibliography{references}'. Whatever you do, DO NOT put the file name extension .bib inside the \bibliography command; this will trip up LaTeX compilers. 
\bibliography{mybib}

%\input{appendix_guidelines}

% \documentclass[runningheads,anonymous]{llncs}
% \usepackage[T1]{fontenc}
% \usepackage{graphicx}
% \usepackage{booktabs}
% \usepackage[misc]{ifsym}
% \usepackage{xcolor}
% \usepackage{amsmath}
% \usepackage{diagbox}
% \usepackage{subcaption}
% \usepackage[hidelinks]{hyperref}
% \newcommand{\corr}{(\Letter)}
% % N.B.: do not change anything above this line. If you require additional packages, please load them directly after this line.
% \usepackage{amsmath, amssymb} %
% \usepackage{multirow}
% \usepackage{tabularx}
% \usepackage[table]{xcolor}

% \begin{document}

% Appendix: Operational Guidelines for G-XAI Practitioners
% To be included in the main paper with \input{appendix_guidelines}

\appendix

\section{Operational Guidelines for G-XAI Practitioners}
\label{app:guidelines}

This appendix provides actionable guidance for machine learning practitioners seeking to integrate explainability into Graph Neural Network pipelines. The recommendations derive from our extensive benchmarking study and are structured around the three phases of ML deployment: pre-development consultation, development-phase validation, and production integration.

\subsection{Explainer Selection Guidelines}
\label{app:explainer_selection}

\subsubsection{Primary Recommendation}

Based on our benchmarking results across four Graph ML tasks, we provide the following explainer selection hierarchy:

\begin{enumerate}
    \item \textbf{Input$\times$Gradient} (Recommended Default): Consistently exhibits excellent performance across all tasks, characterized by high pertinence ($>0.78$), low effective compactness, perfect stability (1.0), and minimal execution time ($<0.03$s). This method should be the default choice unless specific requirements dictate otherwise.

    \item \textbf{Integrated Gradients}: Offers marginally higher pertinence in some scenarios but requires approximately two orders of magnitude more computation time ($0.3$--$1.3$s). Recommended only when explanation quality is essential, and latency constraints are relaxed.

    \item \textbf{Deconvolution / Guided Backpropagation}: Viable alternatives when the primary focus is edge attribution. These methods achieve high edge pertinence ($>0.81$ for classification tasks) with negligible computational overhead.

    \item \textbf{GNNExplainer / LRP}: Acceptable for feature-focused explanations in node classification tasks. However, GNNExplainer exhibits stability issues (0.96--0.97) that may be problematic for regulatory compliance scenarios.
\end{enumerate}

\textbf{Methods to Avoid}: GraphMask consistently yields inadequate results across our benchmarking tasks (pertinence often below 0.5, high effective compactness, low stability) and should be excluded from initial explainer selection.

\subsubsection{Task-Specific Considerations}

Table~\ref{tab:task_recommendations} summarizes explainer recommendations stratified by Graph ML task.

\begin{table}[h]
\centering
\caption{Task-specific explainer recommendations based on benchmarking results.}
\label{tab:task_recommendations}
\small
\begin{tabular}{@{}lll@{}}
\toprule
\textbf{Task} & \textbf{Primary Choice} & \textbf{Alternatives} \\
\midrule
Node Classification & Input$\times$Gradient & IntGrad, LRP (features) \\
Node Regression & Input$\times$Gradient & IntGrad, Deconv (edges) \\
Link Classification & Input$\times$Gradient & IntGrad, GuidedBP (edges) \\
Link Regression & Input$\times$Gradient & IntGrad, Deconv (edges) \\
\bottomrule
\end{tabular}
\end{table}

\subsection{Quality Thresholds and Acceptance Criteria}
\label{app:thresholds}

Establishing appropriate quality thresholds is critical for determining whether an explainer meets deployment requirements. We provide empirically-derived thresholds based on our benchmarking study, while emphasizing that these values should be adapted to specific application domains.

\subsubsection{Stability Thresholds}

Stability values range from 0 (completely random explanations) to 1 (perfectly deterministic). We recommend the following interpretation:

\begin{itemize}
    \item \textbf{$\geq 0.95$}: Highly deterministic; suitable for regulatory reporting where reproducibility is paramount.
    \item \textbf{$0.80$--$0.95$}: Moderate stability; acceptable for internal model diagnostics and development workflows.
    \item \textbf{$< 0.80$}: Potentially problematic; may undermine user trust and should trigger investigation into explainer configuration or selection of alternative methods.
\end{itemize}

For high-stakes applications (e.g., financial fraud detection, anti-money laundering monitoring) where explanations may be scrutinized by auditors or used in legal proceedings, stability thresholds should be set conservatively at $0.80$ or higher.

\subsubsection{Pertinence Thresholds}

Pertinence measures how well explanations capture the model's actual decision-making process compared to random baselines. The metric inherently centers at $0.5$ (equivalent to random):

\begin{itemize}
    \item \textbf{$< 0.5$}: Explanation performs worse than random; should be rejected.
    \item \textbf{$0.5$--$0.6$}: Marginally better than random; acceptable only for inherently difficult tasks.
    \item \textbf{$0.6$--$0.7$}: Moderate alignment with model behavior; acceptable for most applications.
    \item \textbf{$> 0.7$}: Strong alignment; target threshold for classification tasks with balanced classes.
\end{itemize}

\textbf{Important Caveat}: Certain task-data combinations legitimately produce lower pertinence values. In particular, the metric relies on medoid reference targets: when an opposing class is multimodal (several distinct clusters), a single medoid poorly represents the target's nearest alternative, so substitutions follow an incoherent direction and the attribution loses its advantage over random orderings. Practitioners should interpret pertinence in context rather than applying universal thresholds.

\subsubsection{Effective Compactness Thresholds}

Effective compactness quantifies the cognitive load required to comprehend an explanation. For graph data, we cap this metric at 100 for computational tractability. Acceptable ranges depend on dataset characteristics:

\begin{itemize}
    \item \textbf{Feature EC}: We recommend targeting $10$--$25\%$ of total features, depending on task complexity. For node-level tasks, $10\%$ of features represents a reasonable upper bound; for link-level tasks, $25\%$ may be acceptable given the dual-endpoint nature of explanations.

    \item \textbf{Edge EC}: Thresholds are highly dataset-dependent due to varying graph densities and node degrees. We do not prescribe universal thresholds but recommend establishing baselines on representative samples from the target deployment distribution.
\end{itemize}

\subsubsection{Execution Time Thresholds}

Computational efficiency requirements are deployment-dependent:

\begin{itemize}
    \item \textbf{Real-time systems} (interactive dashboards): $< 100$--$300$ ms per explanation.
    \item \textbf{Batch processing} (historical analysis): seconds to minutes per explanation.
    \item \textbf{Offline audit/compliance}: hours of computation acceptable if quality justifies cost.
\end{itemize}

\subsubsection{Summary of Recommended Thresholds}

Table~\ref{tab:thresholds} synthesizes our recommended quality thresholds for graph tasks.

\begin{table}[h]
\centering
\caption{Recommended quality thresholds for graph explainability metrics.}
\label{tab:thresholds}
\small
\begin{tabular}{@{}lccccc@{}}
\toprule
\textbf{Metric} & \textbf{Node Cls.} & \textbf{Node Reg.} & \textbf{Link Cls.} & \textbf{Link Reg.} \\
\midrule
Feature EC & $\leq 10\%$ feats & $\leq 10\%$ feats & $\leq 25\%$ feats & $\leq 25\%$ feats \\
Edge EC & dataset-dep. & dataset-dep. & dataset-dep. & dataset-dep. \\
Feature Pertinence & $\geq 0.6$ & $\geq 0.6$ & $\geq 0.6$ & $\geq 0.6$ \\
Edge Pertinence & $\geq 0.6$ & $\geq 0.6$ & $\geq 0.6$ & $\geq 0.6$ \\
Feature Stability & $\geq 0.8$ & $\geq 0.8$ & $\geq 0.8$ & $\geq 0.8$ \\
Edge Stability & $\geq 0.8$ & $\geq 0.8$ & $\geq 0.8$ & $\geq 0.8$ \\
Execution Time & $< 5$s & $< 5$s & $< 5$s & $< 5$s \\
\bottomrule
\end{tabular}
\end{table}

\subsection{Metric Interpretation and Transferability}
\label{app:interpretation}

\subsubsection{Local vs. Global Assessment}

All four evaluation metrics are computed at the instance level (individual nodes or edges) and subsequently aggregated. This design reflects that explanation quality varies across different regions of the input space. To obtain statistically reliable global assessments:

\begin{itemize}
    \item \textbf{Minimum sample size}: 50--100 instances for initial assessment.
    \item \textbf{Robust validation}: 200--500 instances for production deployment.
    \item \textbf{Stratified sampling}: When computational budgets are constrained, stratify across prediction confidence levels and/or classes to capture explainer behavior in both high-certainty and ambiguous decision regions.
\end{itemize}

\subsubsection{Transferability from Benchmark to Production}

Different metrics exhibit varying degrees of transferability:

\begin{itemize}
    \item \textbf{Stability} (High Transferability): Primarily depends on algorithmic determinism rather than data characteristics. A stochastic explainer producing inconsistent results on benchmark graphs will likely exhibit similar instability on proprietary data.

    \item \textbf{Execution Time} (High Transferability): Scales proportionally with graph complexity (nodes, edges, layers). Benchmark timings provide reliable estimates when adjusted for hardware differences.

    \item \textbf{Pertinence} (Medium Transferability): Sensitive to data distribution, feature correlations, and decision boundary complexity. Requires domain-specific validation.

    \item \textbf{Effective Compactness} (Medium Transferability): Affected by number of features, graph density, and model architecture (e.g., transitioning from 1-hop to 3-hop GNNs significantly increases computational graph size).
\end{itemize}

Practitioners should interpret benchmarking results as informative priors rather than absolute guarantees for novel problem settings.

\subsection{Computational Optimization Strategies}
\label{app:optimization}

\subsubsection{Early Stopping for Iterative Metrics}

Both stability and pertinence require iterative computation (100 trials by default). We implement an early stopping criterion that reduces computation by 60--80\% with minimal accuracy loss:

\begin{enumerate}
    \item After 30 trials, compute running mean $\mu$ and standard error $\text{SE} = \sigma / \sqrt{n}$.
    \item \textbf{Mean convergence test}: Terminate if $z = |\mu - \tau| / \text{SE} > 1.645$ (one-tailed, 95\% confidence), where $\tau$ is an optional threshold.
    \item \textbf{Precision tolerance test}: Terminate if $z = \epsilon / \text{SE} > 1.96$ (two-tailed, 95\% confidence), where $\epsilon$ is the acceptable tolerance (default: 0.05).
\end{enumerate}

This approach reduced average iterations from 100 to 35--40 in our benchmarking study.

\subsubsection{Binomial Algorithm for Stability}

For deployment scenarios requiring rapid stability assessment, we provide an alternative ``binomial'' algorithm:

\begin{enumerate}
    \item Extract $M$ explanations such that $M(M-1)/2 \approx N$ (target number of pairs).
    \item Compute stability across all $\binom{M}{2}$ possible pairs.
    \item Return the average pairwise stability.
\end{enumerate}

This approach requires only $M$ explanation generations instead of $2N$, significantly reducing computational overhead for production monitoring.

\subsubsection{Trade-off Analysis}

We recommend conducting parallel experiments during development:

\begin{itemize}
    \item \textbf{Exhaustive evaluation}: Full 100-iteration metric computation on development set.
    \item \textbf{Early stopping evaluation}: Same computation with early stopping enabled.
    \item \textbf{Comparison}: Quantify precision-efficiency trade-off to inform production configuration.
\end{itemize}

The acceptable trade-off depends critically on inference pipeline architecture: real-time systems with sub-second latency constraints necessitate aggressive approximations, whereas batch pipelines executed during off-peak hours can afford exhaustive computation.

\subsection{Deployment Workflow}
\label{app:workflow}

\subsubsection{Phase 1: Pre-Development Consultation}

Before model development, leverage benchmarking results to inform explainability strategy:

\begin{enumerate}
    \item Identify task type (node/link classification/regression).
    \item Consult benchmark rankings (Table~\ref{tab:task_recommendations}) to shortlist candidate explainers.
    \item Evaluate trade-offs: pertinence vs. execution time, feature vs. edge attribution quality.
    \item Document explainability requirements aligned with regulatory obligations (EU AI Act, domain-specific mandates).
\end{enumerate}

\subsubsection{Phase 2: Development Validation}

During model development, validate that benchmark performance transfers to proprietary data:

\begin{enumerate}
    \item Wrap trained model using appropriate model class (e.g., \texttt{NodeClassificationModel}).
    \item Configure \texttt{Evaluator} with dataset-specific parameters.
    \item Evaluate shortlisted explainers on representative sample ($\geq 100$ instances).
    \item Compare metrics against thresholds (Table~\ref{tab:thresholds}).
    \item Select final explainer based on validated performance.
\end{enumerate}

\subsubsection{Phase 3: Production Integration}

Deploy selected explainer alongside ML model:

\begin{enumerate}
    \item Integrate explainer into inference pipeline.
    \item Configure natural language translation for end-user consumption.
    \item Establish monitoring protocols to track metric stability over time.
    \item Define re-validation triggers: model retraining, data distribution shift, regulatory updates.
\end{enumerate}

\subsection{Translating Explanations for Regulatory Compliance}
\label{app:nlp}

Technical explanations (attribution matrices, edge importance vectors) require translation for non-technical stakeholders. Our framework provides natural language generation capabilities that convert low-level explainer outputs into narrative explanations.

%\subsubsection{Attribution-based Explanations}

For attribution methods, the translation follows this template:

\begin{quote}
\textit{``The model's prediction for [target] was primarily influenced by [top-k features] of [target node/nodes]. The structural context, particularly connections to [important neighbors], also contributed to the decision. Specifically, [feature 1] had the highest impact, followed by [feature 2] and [feature 3].''}
\end{quote}

\subsubsection{Regulatory Alignment}

These natural language outputs address key regulatory requirements:

\begin{itemize}
    \item \textbf{Transparency}: Explanations are human-readable without technical expertise.
    \item \textbf{Accountability}: Specific features and connections are identified, enabling audit trails.
    \item \textbf{Contestability}: Affected individuals can understand and challenge decisions based on the explanation.
\end{itemize}

\subsection{Common Pitfalls and Troubleshooting}
\label{app:troubleshooting}

Table~\ref{tab:troubleshooting} summarizes common issues encountered during G-XAI deployment and recommended resolutions.

\begin{table}[h]
\centering
\caption{Troubleshooting guide for common G-XAI deployment issues.}
\label{tab:troubleshooting}
\small
\renewcommand{\arraystretch}{1.5}
\begin{tabularx}{\textwidth}{@{}p{3cm}|X|X@{}}
\toprule
\textbf{Issue} & \textbf{Probable Cause} & \textbf{Resolution} \\
\midrule
Pertinence $< 0.5$ & Inherently difficult task (anomaly detection on normal points, regression near baseline) & Verify if legitimate for task; consider alternative evaluation approaches \\
\hline
Low stability & Stochastic explainer (GNNExplainer, GraphMask) & Switch to deterministic method (Input$\times$Gradient) \\
\hline
High effective compactness & Dense graphs, many features & Accept if unavoidable; consider hierarchical explanation presentation \\
\hline
Excessive execution time & Using IntGrad, GNNExplainer, or GraphMask & Switch to Input$\times$Gradient or other gradient-based methods \\
\hline
Unstable metrics across samples & Insufficient sample size & Increase evaluation set to 200+ instances \\
\hline
Metrics don't match benchmark & Distribution shift, different model architecture & Re-validate on proprietary data; adjust thresholds accordingly \\
\hline
NaN/invalid attributions & Numerical instability, incompatible model & Check model compatibility; verify preprocessing pipeline \\
\bottomrule
\end{tabularx}
\end{table}

\subsection{Checklist for Production Deployment}
\label{app:checklist}

Before deploying G-XAI in production, verify completion of the following:

\begin{enumerate}
    \item[$\square$] Explainer selected based on benchmark consultation
    \item[$\square$] Validated on proprietary data ($\geq 100$ representative samples)
    \item[$\square$] All metrics meet or exceed recommended thresholds
    \item[$\square$] Early stopping / optimization configured if latency-constrained
    \item[$\square$] Natural language translation verified for target audience
    \item[$\square$] Monitoring dashboard configured for ongoing metric tracking
    \item[$\square$] Re-validation protocol documented (triggers, frequency)
    \item[$\square$] Regulatory compliance documentation prepared
\end{enumerate}

\section{Datasets Details}

In this Section we provide additional details on each of the datasets which we used in our experiments, providing more details for each of them. In Table \ref{tab:datasets} we summarise their statistics.

\begin{table}[h]
    \centering
    \caption{Summary of dataset statistics. "Reg." is used instead of a number of classes in datasets with regression tasks.}
    \begin{tabularx}{\textwidth}{c|XXXXX}
    \toprule
        Dataset & Num. \newline Nodes & Num. \newline Edges & Num. \newline Features & Num.\newline Classes & Num. \newline Edge Classes \\
        \midrule
        Cora & 2,708 & 8,448 & 1,433 & 7 & 2 \\
        GitHub & 37,700 &462,406 & 128 &2 & 2 \\
        BA-Shapes & 700 & 3,942 & 700 & 4 & 6\\
        Movielens & 1,180 & 226,272 & 404 & Reg. & Reg. \\
        Synthetic & 10,000 & 150,324 & 16 & Reg. & Reg.\\
        \bottomrule
    \end{tabularx}
    \label{tab:datasets}
\end{table}

\subsubsection{Cora} The Cora dataset is a standard benchmark dataset for node classification and link prediction. represents a citation network, where each node is an article and an edge implies a citation between papers. The graph is undirected, since the effective direction of a citation is not taken into account. The dataset has 2,708 nodes and 4,224 edges. Each node has a bag-of-words as features, with 1,433 different terms.  The node task is to correctly classify a node according to the seven subjects of the datasets, while the link task is to predict if a given citation exists or not.

\subsubsection{GitHub} The GitHub dataset is a larger and denser benchmark, connecting approximately 37,700 nodes representing developers through 231,203 "follows". This dataset allows for the creation of larger explanations to test. It supports a node-based classification task to distinguish between web and machine learning developers, and an edge-based task to predict the existence of a "follow" relationship between developers. The features are a 128 dimension embedding for each node.

\subsubsection{BAShapes} Another common benchmark, specifically used to test explainers. The dataset is composed of 100 small synthetic graphs of 70 nodes each, produced by a Barabási-Albert model, each of which has a small, “house-shaped” structure, strongly distinct from the rest of the graph in terms of connectivity. Each node is initalized with a one-hot encoding embedding, to distinguish it from the others inside the GNN, but they have no feature otherwise. The task on the nodes is to predict if a node belongs to the random graph, or if it belongs to the “base”, the “top” or the “roof” of the “house”. Similarly, on edges we have defined a multiclassification problem, where the task is to predict which part of the motif the edge is (e.g., part of the random graph, connecting the random graph to the “house”, or a specific edge of the “house” motif).

\subsubsection{Movielens} A dataset extracted from a larger and richer dataset from the website “MovieLens”, collecting movies and users who review them. Out Movielens dataset is a graph of 1,180 movies which are connected if they share at least 20 watchers who reviewed them. It is very dense, with 113,136 edges, and each node has a feature vector of 404 features representing the genres to which they can belong. On the nodes, we defined a regression task to predict the average movie rating, while on the edges we must predict the fraction of watchers which two movie share. When performing the regression on the nodes, we use the shared fraction of watchers as additional edge weights. 

\subsubsection{Synthetic} The last dataset is a full synthetic graph. The graph structure has been generated from a Erdős-Rényi model, with 10,000 nodes and 75,162 edges ($p=0.0015$ that two nodes have an edge). Each edge has a set of 16 random features, generated from a normal distribution of mean 0 and variance 1. Both on nodes and edges we define a custom regression task. The node labels have been produced by an arbitrary, weighted aggregation between the features of a node and its 2-hop neighborhood (i.e. all its neighbours and the neighbours of its neighbours). The aggregation gives different weights to each of the original features, making certain features more relevant than others.  The edge labels are instead given by sums between the node-level labels of the edge vertices. In both cases, the purpose is to retrieve the original arbitrary aggregation rule (or an approximation).

\section{Models}

As mentioned in the main text, we used two different model architectures: a 2-layer GCN and a 2-layer GAT. We trained all models for 200 epochs, an early stopping of 10 epochs, dropout regularization, and Adam optimizer. For each dataset and task, we performed an hyperparameter grid search with the following hyperparameters and ranges:
\begin{itemize}
    \item Hidden channels in $[8, 16,32, 64,128]$;
    \item Learning rate in $[10^{-4}, 10^{-3},10^{-2}]$;
    \item Weight decay in $[10^{-5},10^{-4},10^{-3}]$;
    \item Dropout probability in $[0.0, 0.2, 0.3, 0.4,0.5]$.
\end{itemize}

Classification models use a ReLU non-linearity between the two layers, while regression models use a LeakyReLU. Additionally, models for link task use an additional decoder to perform edge-level predictions, which consist of a inner product between the embeddings (with size given by the hidden channels) of the vertices of the target edge. For the edge-level task on the Synthetic dataset, we performed a further simplification, replacing the inner product with a sum between the two (scalar) embeddings of the vertices.

\begin{table}[h!]
    \centering
    \caption{Model performances on classification tasks, measured as accuracy.}
    \begin{tabularx}{0.6\textwidth}{l|cc|cc}
    \toprule
    Task  & \multicolumn{2}{c|}{Node Classification} &  \multicolumn{2}{c}{Link Classification} \\
    Model & GCN & GAT & GCN & GAT \\
    \midrule
    BA-Shapes & 0.81 & 0.87 & 0.97 & 0.99 \\
    Cora & 0.76 & 0.79 & 0.86 & 0.92 \\
    Github & 0.86 & 0.86 & 0.62 & 0.60 \\
    \bottomrule
    \end{tabularx}
    \label{tab:classification_performances}
\end{table}

We report all the results in Table~\ref{tab:classification_performances} and Table~\ref{tab:regression_performances}. Classification tasks are evaluated through accuracy, while we use $r^2$ for regression tasks. Although the GCN model underperforms on both link regression on Sythetic and node regression on Movielens, we do not expect this to impact the pertinence or the effective compactness of the explanations, as these metrics do not take into account the true labeling of a target, but only the model rationale in predicting a label. As long as the model's output is not a constant value (which would remove the existence of reference targets with alternative predictions, fundamental for our algorithms), the metrics can still be measured correctly.

\begin{table}[h!]
    \centering
    \caption{Model Performances measured as $r^2$ in regression tasks.}
    \begin{tabularx}{0.52\textwidth}{l|cc|cc}
    \toprule
    Task  & \multicolumn{2}{c|}{Node Regression} & \multicolumn{2}{c}{Link Regression} \\
    Model & GCN & GAT & GCN & GAT \\
    \midrule
    Movielens  & 0.44 & 0.64 & 0.78 & 0.76 \\
    Synthetic & 0.80 & 0.90 & 0.31 & 0.64 \\
    \bottomrule
    \end{tabularx}
    \label{tab:regression_performances}
\end{table}

\section{Hardware Architecture}
    All experiments were executed on a machine equipped with an NVIDIA A100 GPU (80GB VRAM) and an Intel Xeon Gold 6248R 15-core CPU.

\end{document}